\documentclass[conference]{IEEEtran}
\IEEEoverridecommandlockouts
\usepackage{cite}
\usepackage[colorlinks=true, linkcolor=black, urlcolor=black, citecolor=black]{hyperref}
\usepackage{braket}
\usepackage{booktabs}
\usepackage{threeparttable}
\usepackage{multirow} 
\usepackage{makecell}
\usepackage{pifont}

\usepackage[caption=false,font=footnotesize,labelfont=sf,textfont=sf]{subfig}
\usepackage{textcomp}
\usepackage{tabularx}
\usepackage{amsmath,amssymb,amsfonts}
\usepackage{algorithmic}
\usepackage{graphicx}
\usepackage{marvosym}
\usepackage{textcomp}
\usepackage{xcolor}
\newcommand{\hhline}{\noalign{\vskip 1pt}\hline\noalign{\vskip 1pt}}
\newcommand{\slfrac}[2]{\left.#1\middle/#2\right.}
\newcommand{\degree}{^\circ}
\newcommand{\etal}{et al.}
\def\BibTeX{{\rm B\kern-.05em{\sc i\kern-.025em b}\kern-.08em
    T\kern-.1667em\lower.7ex\hbox{E}\kern-.125emX}}
\begin{document}

\title{Fixed-length Dense Descriptor for Efficient Fingerprint Matching
\thanks{\Letter  Jianjiang Feng is the corresponding author. \\ \indent This work was supported in part by the National Natural Science Foundation of China under Grant 62376132 and 62321005.}
}

\author{
\IEEEauthorblockN{Zhiyu Pan, Yongjie Duan, Jianjiang Feng\textsuperscript{\Letter} and Jie Zhou}
\IEEEauthorblockA{\textit{Department of Automation, BNRist, Tsinghua University}, Beijing, China \\
pzy20@mails.tsinghua.edu.cn, duanyj13@tsinghua.org.cn,  \{jfeng, jzhou\}@tsinghua.edu.cn}
}

\maketitle
\begin{abstract}
    In fingerprint matching, fixed-length descriptors generally offer greater efficiency compared to minutiae set, but the recognition accuracy is not as good as that of the latter. Although much progress has been made in deep learning based fixed-length descriptors recently, they often fall short when dealing with incomplete or partial fingerprints, diverse fingerprint poses, and significant background noise. In this paper, we propose a three-dimensional representation called Fixed-length Dense Descriptor (FDD) for efficient fingerprint matching. FDD features great spatial properties, enabling it to capture the spatial relationships of the original fingerprints, thereby enhancing interpretability and robustness. Our experiments on various fingerprint datasets reveal that FDD outperforms other fixed-length descriptors, especially in matching fingerprints of different areas, cross-modal fingerprint matching, and fingerprint matching with background noise.
\end{abstract}
\begin{IEEEkeywords}
Fingerprint Matching, Fixed-length Dense Descriptors, Deep Learning.
\end{IEEEkeywords}

\section{Introduction}
Fingerprint recognition has become a prevalent biometric technology used in various applications, such as access control, national ID cards, and law enforcement \cite{maltoni2022handbook}. Typically, a fingerprint recognition system comprises three core modules: image acquisition, feature extraction, and matching. The feature extraction algorithm is designed to extract distinctive features, known as descriptors, from fingerprints. The matching algorithm then uses these descriptors to compute a matching score between samples. Numerous researchers have focused on developing automated and efficient algorithms for descriptor extraction and matching \cite{cappelli2010minutia, song2017fingerprint, cao2019automated, grosz2023latent}, aiming to meet the diverse demands of various fingerprint recognition applications.

Generally, fingerprint representations are categorized into two main types: minutiae set and fixed-length descriptors. Minutiae set based representations describe fingerprints by focusing on individual local minutiae, with each minutia having its associated descriptor. Most local minutiae descriptors \cite{cappelli2010minutia, cao2019automated, pan2024latent} are generated by extracting patch images centered around the location and orientation of each minutia. This method provides a flexible representation of fingerprints by capturing detailed local features in the ridges and valleys, which enhances accuracy and adaptability to various fingerprint acquisition techniques and conditions. However, minutiae-based fingerprint representations pose several challenges, including high computational complexity \cite{cappelli2010minutia, cao2019automated}, difficulties in template encryption \cite{rane2013secure}, and dependence on the precision of minutiae extraction \cite{sankaran2014latent, tang2017fingernet}.

Fixed-length descriptors represent each fingerprint as a fixed-length tensor, allowing the similarity between multiple fingerprints to be efficiently calculated through parallel matrix multiplication. In the past, conventional fixed-length descriptors were typically defined by the local responses of multiple Gabor filters \cite{jain1999fingercode}, parameters of the orientation field model \cite{wang2007fingerprint, Kumar2011}, local minutiae vicinities \cite{bringer2010binary}, histograms of minutiae triplet or pair features \cite{farooq2007anonymous, jin2010revocable}, local ridge orientation and frequency \cite{cappelli2011fast}, and local discretized texture descriptors \cite{imamverdiyev2013biometric}. These handcrafted representations often exhibited limited discriminating capability, leading to insufficient matching accuracy.

\begin{figure}[!t]
    \centering
    \includegraphics[width=\linewidth]{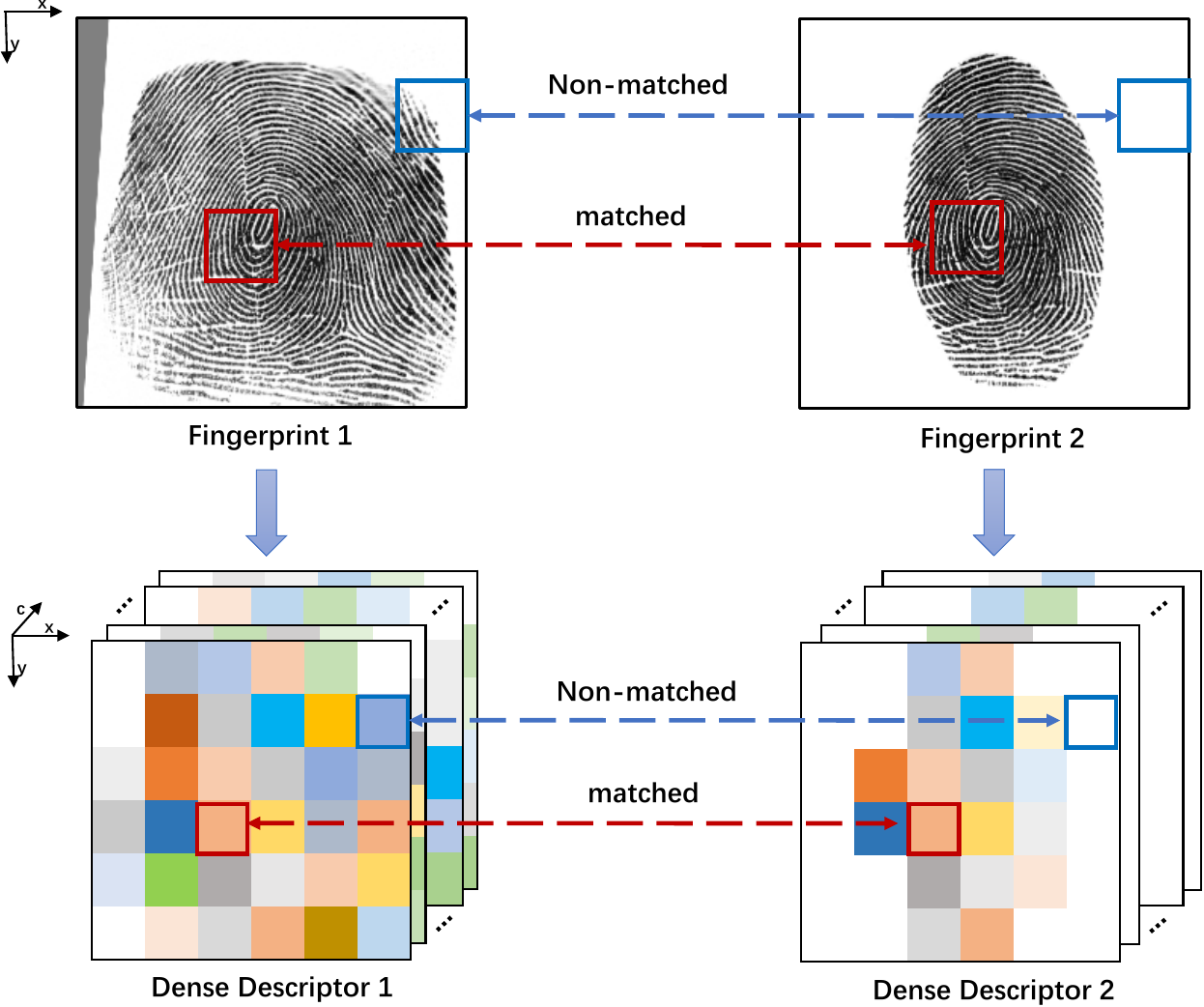} 
    \caption{The illustration of Fixed-length Dense Descriptors (FDD) of a rolled fingerprint and a mated plain fingerprint. Two corresponding regions and their features are marked.}
    \label{fig:FDD_illus}
    \vspace{-0.5cm}
\end{figure}

With the rapid development of deep learning technology, fixed-length representations are now extracted using deep networks from aligned fingerprints \cite{cao2017fingerprint, engelsma2021learning, grosz2023afrnet}. These methods demonstrate exceptional performance on high-quality large fingerprints, such as rolled fingerprints. However, most existing techniques represent a fingerprint image using a one-dimensional descriptor, which fails to sufficiently account for background areas. This limitation significantly decreases matching performance, especially when dealing with partial fingerprints or significant background noise. Furthermore, the matching of fixed-length one-dimensional descriptors is not restricted to the overlapping foreground area, reducing their effectiveness for fingerprints with varying areas. Gu et al. \cite{gu2022latent} explored extracting pyramid features at different scales and positions. Despite this approach, matching performance remains hindered by the coarse division of local regions and the fusion weights for different fingerprints. Additionally, extracting these multi-scale features demands higher computational resources and storage space. 

Given the strong performance of dense descriptors in characterizing neighborhood of minutiae \cite{pan2024latent}, this paper extends the concept to the whole fingerprint and introduces the Fixed-length Dense Descriptor (FDD). Dense descriptors are three-dimensional representations, where two dimensions correspond to the spatial dimensions of the original fingerprint image (Fig. \ref{fig:FDD_illus}). Consequently, dense descriptors contain valid values only within the fingerprint foreground, effectively eliminating the background. Additionally, the spatial alignment of fingerprints to be matched ensures that the dense descriptor spaces are also aligned, thus restricting the matching process to the overlapping foreground regions.

Accordingly, the first step in our proposed method is to align the fingerprint using a robust fingerprint pose estimation approach \cite{duan2023estimating}. Next, we extract the localized dense description of the fingerprint and identify the valid area. To add global constraints on the aligned space, we utilize a 2D sinusoidal positional embedding, introducing position information into the fixed-length representations. This guides the model to distinguish similar fingerprint textures at different locations, further enhancing matching performance. 

To verify the effectiveness and generalization ability of FDD, we evaluate it on diverse datasets containing various fingerprint areas and impression types, including rolled, plain, latent, and contactless fingerprints. Some of these datasets present significant challenges for fixed-length descriptors, such as fingerprints with diverse finger poses and small image sizes. The experimental results demonstrate the superiority of FDD compared to other fixed-length descriptors. Additionally, we conduct experiments on binarized FDDs, obtained using a constant threshold, which still achieve good performance. Therefore, the Fixed-length Dense Descriptor shows great potential for application in automated fingerprint recognition systems.

\section{Related Work} \label{sec:related}
Fixed-length fingerprint descriptor can be classified into two categories: alignment-based and alignment-free.

Currently, numerous high-performance and robust fixed-length descriptors leverage deep learning frameworks to extract more discriminative features.
In alignment-based methods, fingerprints are initially aligned, followed by the extraction of fixed-length descriptors from these aligned images. Cao et al. \cite{cao2017fingerprint} introduced the extraction of deep features from aligned fingerprints, while Engelsma et al. \cite{engelsma2021learning} developed an end-to-end network integrating alignment and feature extraction. Grosz et al. \cite{grosz2023afrnet} improved recognition by fusing attention-based and CNN-based embeddings. These techniques excel with large, complete fingerprint areas, like rolled prints, but may be compromised by background noise in incomplete prints, affecting matching accuracy. Song and Feng\cite{song2017fingerprint} and Gu et al.\cite{gu2022latent} proposed using pyramid features for global and local representations to address incomplete fingerprints, but this approach still has issues with region overlap and requires dataset-specific optimizations. Some researchers \cite{PFVNet, qiu2024ifvit} boost accuracy through concurrent alignment and scoring of two images, which, however, increases computational demands due to the dual-image input costs.

Alignment-free fingerprint descriptors typically utilize minutiae data. Song et al. \cite{song2019aggregating} and Wu et al. \cite{wu2022minutiae} aggregate local features around minutiae into global descriptors, with the latter using multi-task learning. Li et al. \cite{li2019learning} integrate minutiae via a fully convolutional network and global average pooling. However, these methods may be affected by minutiae extraction variability, and the loss of minutiae count and positions post-aggregation diminishes feature discriminability.

Fixed-length Dense Descriptor proposed in this paper is alignment-based. To alleviate the negative impact of inaccurate fingerprint alignment, a fingerprint pose estimation algorithm, which achieves state-of-the-art performance across various fingerprint datasets \cite{duan2023estimating}, was employed to align fingerprints.

\begin{figure}[!t]
    \centering
    \includegraphics[width=\linewidth]{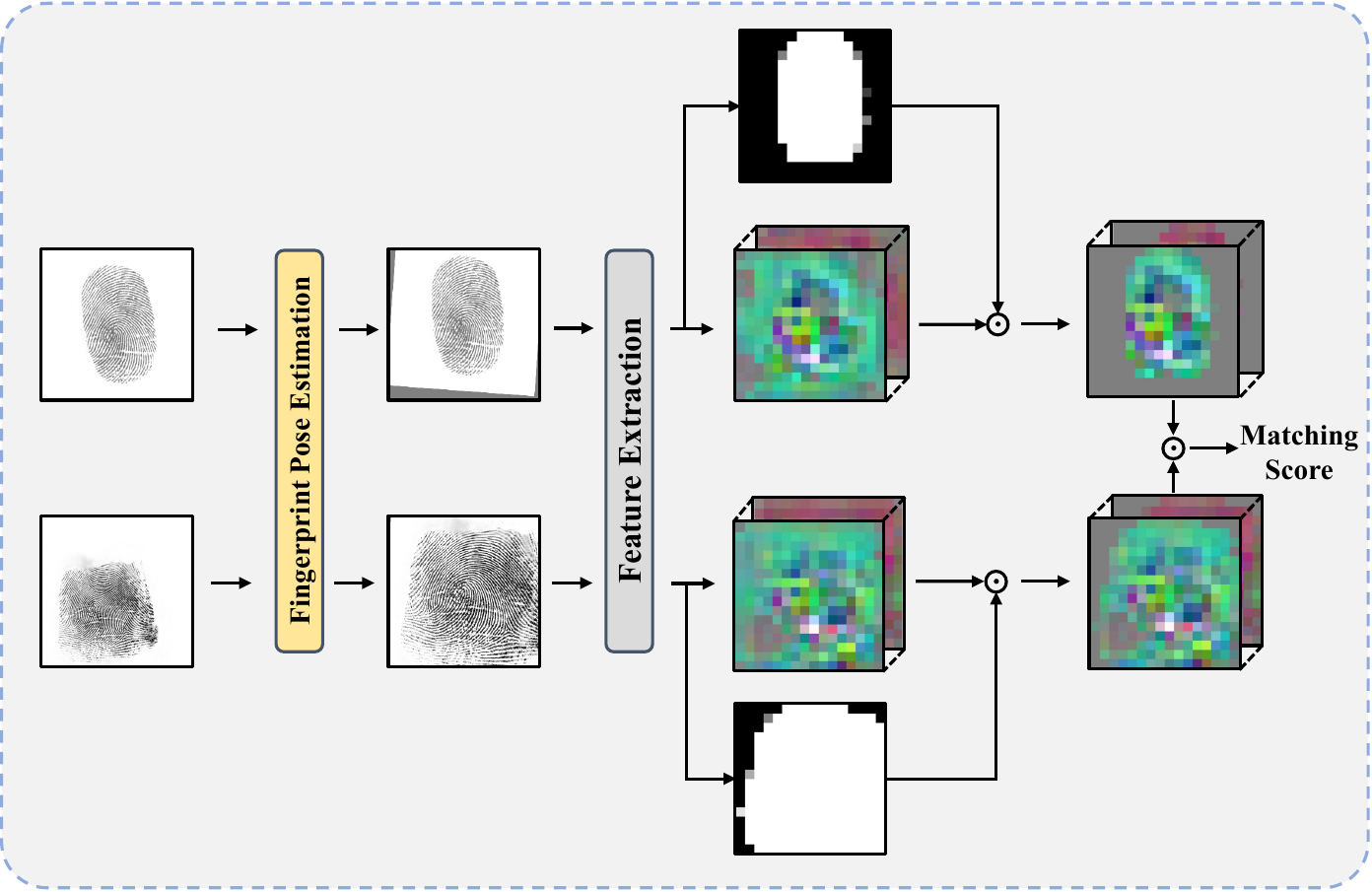}
    \caption{Overview of fingerprint matching using the proposed FDD. First, fingerprint pose alignment is performed according to Duan et al.\cite{duan2023estimating}, followed by the extraction of fixed-length dense descriptors with the descriptor extraction network, which are then used for matching.}
    \label{fig:fp_matching}
    \vspace{-0.3cm}
\end{figure}

\section{Method}
We propose the Fixed-length Dense Descriptor (FDD) for fingerprint matching with efficiency and interpretability. Fingerprints are first aligned using the estimated fingerprint poses \cite{duan2023estimating}. Subsequently, localized representations are extracted, and the segmentation mask is determined simultaneously. The final FDD is obtained by element-wise multiplying the local representations with the predicted mask. And the matching score is computed based on the cosine similarity of two normalized flattened FDD. The overall schematic illustration of fingerprint matching using FDD is shown in Fig. \ref{fig:fp_matching}.
\subsection{Network Design}
\begin{figure}[!t]
    \centering
    \includegraphics[width=\linewidth]{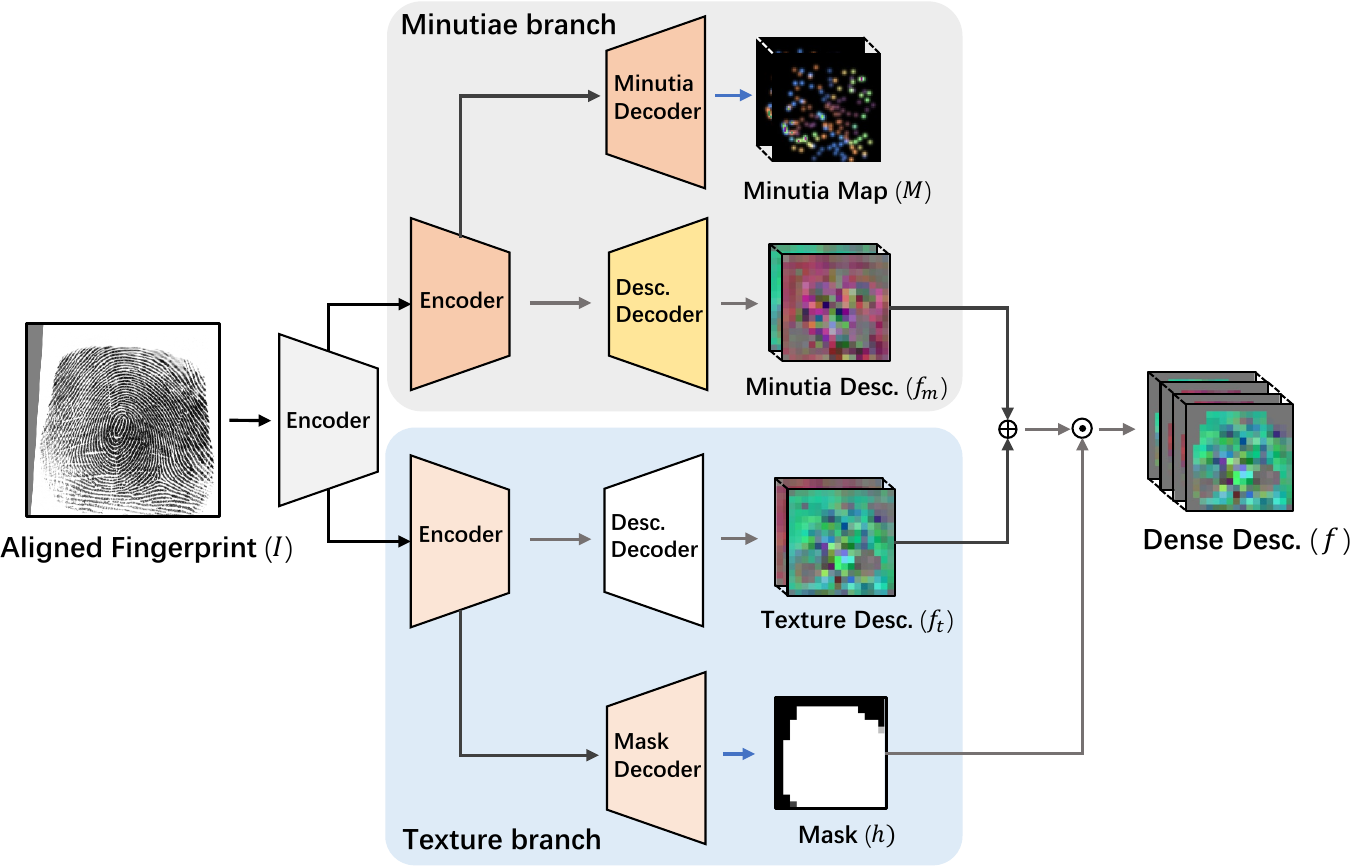}
    \caption{The overall network architecture of FDD consists of the minutia branch and the texture branch, which extract minutia maps and masks, respectively, along with their corresponding descriptors. ``Desc.'' is an abbreviation for ``Descriptor''.}    
    \label{fig:network}
    \vspace{-0.3cm}
\end{figure}

We draw upon the method of extracting Dense Minutia Descriptor \cite{pan2024latent}, employing ResNet-34 \cite{he2016deep} as the backbone of our network, and modify it to a dual-branch structure. To retain more details of ridges, we remove the first max pooling layer, resulting in an output size that is $\slfrac{1}{16}$ of the input size. Considering the compromise between computational efficiency and performance, we cropped the aligned fingerprint with 500 ppi to $512\times512$ and subsequently scaled it down to $256\times256$ as input for the network. Therefore, the spatial resolution of FDD in our case is $16 \times 16$. The overall network structure is shown in Fig. \ref{fig:network}.

Except for global semantic information, local positional information is also essential for incorporating global constraints due to the alignment among the fingerprint spaces. We utilize the classic 2D sinusoidal positional embedding module \cite{vaswani2017attention} to enable the network focus on distinct characteristics across different local regions, resulting in a more refined feature representation.

We design a dual-branch structure consisting of a minutiae branch and a texture branch. Each branch follows a multi-task approach, incorporating dense descriptor extraction along with a corresponding auxiliary task. Minutiae branch outputs the minutiae distribution in a heatmap form called minutia map \cite{engelsma2021learning} denoted as $M$. The extraction of the minutia map $M \in \mathbb{R}^{6 \times 128 \times 128}$ and minutia descriptor $f_m \in \mathbb{R}^{C \times 16 \times 16}$ share features from Minutia Feature Encoder. Similarly, for the texture branch, mask $h \in \mathbb{R}^{16 \times 16}$ prediction and texture descriptor $f_t \in \mathbb{R}^{C \times 16 \times 16}$ deriving also share the same features. The final Fixed-length Dense Descriptor $f \in \mathbb{R}^{2C \times 16 \times 16}$ is obtained by concatenating two branch's descriptors and multiplying the mask as 
\begin{equation}
  \label{eq:generate_desc}
   f = (f_\text{t} \oplus f_\text{m}) \odot h,
\end{equation}
where $\oplus$ denotes the concatenation and $\odot$ denotes the Hadamard product. 

\subsection{Fingerprint Matching}
Given a pair of fingerprints $(p,g)$, the extracted FDDs by Eq. \eqref{eq:generate_desc} are denoted as $f^q\in\mathbb{R}^{2C\times16\times16}$ and $f^g\in\mathbb{R}^{2C\times16\times16}$, and the estimated mask area as $h^q\in\mathbb{R}^{1\times16\times16}$ and $h^g\in\mathbb{R}^{1\times16\times16}$. 
The matching score is obtained by calculating the cosine similarity between the two flattened dense descriptors ${f^q}^\prime$ and ${f^g}^\prime$. Only the feature embeddings within the overlapping area are considered during comparison. Specifically, the matching score of fingerprint pair $(p,g)$ is computed by
\begin{equation}
    s(q,g) = \frac{1}{2} \cdot \frac{{{f^q}^\prime}^\mathrm{T} {f^g}^\prime}{\|f^q\odot h^g\|_F~\|f^g \odot h^q\|_F} + \frac{1}{2},
\end{equation}
where ${f^q}^\prime \in \mathbb{R}^{512C}$, ${f^g}^\prime \in \mathbb{R}^{512C}$, and $\|\cdot\|_F$ represents Frobenius norm.

\subsection{Objective Functions}
The proposed network aims to extract robust and discriminative representations, and the following provides detailed descriptions of the objective functions.
\subsubsection{Classification Loss} Inspired by face recognition deep models, we utilize the CosFace loss \cite{wang2009detecting} to supervise the descriptors' extraction. Specifically, the extracted dense descriptors are first flatten into one dimension and followed by a Fully Connected (FC) layer with $K$ classes. The CosFace loss function is modified based on SoftMax
\begin{equation}\label{eq:cosface}
    \mathcal{L}_\text{cls}^{i} = -\frac{1}{N} \sum_{n=1}^{N} \log \frac{e^{A(\cos(\theta_{y_n}^{i}) - b)}}{e^{A(\cos(\theta_{y_n}^{i}) - b)} + \sum_{k=1,k\neq y_n}^{K}e^{A\cos(\theta_k)}},
\end{equation}
where $\cos(\theta_k)$ is calculated by
\begin{equation}
\cos(\theta_k ^ i) = W_k^\mathrm{T}f_i,\quad W_k=\frac{W_k}{\| W_k \|}, \quad f_i = \frac{f_i}{\| f_i \|}.
\end{equation}
The hyper-parameter $A$ is utilized for magnitude normalization and $i \in [t,m]$ denotes the descriptor's type, $b$ denotes the margin, $N$ is the number of samples within a batch, and $y_n$ is the class label of sample $f_i^n$. The $W_k$ represents the learnable weight vector of the $k$-th class.
\subsubsection{Auxiliary Loss}
The auxiliary tasks contain the minutia map extraction and mask prediction, therefore the auxiliary loss includes the binary cross entropy loss for calculating $\mathcal{L}_{\text{mask}}$ and mean square error for calculating minutia heatmap loss $\mathcal{L}_{\text{mnt}}$. We utilize the VeriFinger v12.0 \footnote{\href{https://neurotechnology.com/verifinger.html}{https://neurotechnology.com/verifinger.html}} to extract the target minutiae and masks.

\subsubsection{Local Similarity Loss} 
Given that $\mathcal{L}_\text{cls}$ considers all elements of the extracted localized representations, the features from incomplete fingerprints cannot be directly optimized. Consequently, we proposed a ``teaching-mimicking'' tactic, wherein features extracted from incomplete fingerprints mimic those found in complete fingerprints to uphold consistency in representations across fingerprints of various visible areas, intensity distributions, and distortion patterns originating from the same finger. We simulate the incomplete fingerprints by adding masks on the complete ones. Specifically, the loss function is defined by
\begin{equation}\label{eq:localsimilarity}
    \mathcal{L}_\text{sim} = \frac{1}{|h_{q\cap g}|} \sum\nolimits_{(i,j)\in h_{q\cap g}} \| f^q(i,j) - f^g(i,j) \|^2,
\end{equation}
where $f_q$ and $f_g$ denote the representations extracted from incomplete and complete fingerprints respectively. 

Consequently, the proposed network was supervised by optimizing the overall objective function
\begin{equation}
  \mathcal{L} = \sum_{i\in\{t,m\}}\mathcal{L}_\text{cls}^i + \lambda_\text{mask}\mathcal{L}_\text{mask} + \lambda_\text{minu}\mathcal{L}_\text{minu} + \lambda_\text{sim}\mathcal{L}_\text{sim},
\end{equation}
where $\lambda_\text{mask}$, $\lambda_\text{minu}$, and $\lambda_\text{sim}$ are trade-off parameters for $L_\text{mask}$, $L_\text{minu}$, and $L_\text{sim}$ respectively. In this paper, we set $\lambda_\text{mask}=1$, $\lambda_\text{minu}=0.01$, and $\lambda_\text{sim}=0.00125$ to achieve roughly the same decreasing rate for those loss functions.

\subsection{Implementation Details}
We designed a data augmentation strategy to simulate multiple impressions from the same finger, thus increasing the diversity of training data to avoid over-fitting. Specifically, we utilized the fingerprint distortion model in Si \etal \cite{si2015detection} and randomly generated different distortion fields, which were then applied on real fingerprints to simulate impressions with different distortion patterns. Histogram matching of intensity was then employed to adjust the intensity distribution of synthetic fingerprints with arbitrary plausible distributions, which aims to increase the variations of intensity distribution in synthetic fingerprints. Moreover, additional data augmentation was also adopted, including random translation in range of 10 pixels and random rotation to follow a uniform distribution $[-5\degree,5\degree]$. The feature dimension $C$ of our Fixed-length Dense Descriptor is set to $6$, therefore $f \in \mathbb{R}^{12 \times 16 \times 16}$. The normalization parameter $A$ and margin $b$ in Eq. \eqref{eq:cosface} are set to 30 and 0.4, respectively. AdamW optimizer with initial learning rate of 0.00035 was utilized for network training. Learning rate decays by 0.1 after 10 epochs without performance improvement on validation dataset. And early stopping was performed after three times learning rate decays. Furthermore, L2 regularization on trainable parameters was also utilized to avoid over-fitting.

\section{Experiments} \label{sec:experiments}
In this section, we first introduce the datasets utilized in this paper and the comparing methods, then the evaluation of matching performance on these datasets. 
\subsection{Datasets}
\begin{figure}[!t]
    \centering
    \subfloat[]{\includegraphics[width=.18\linewidth]{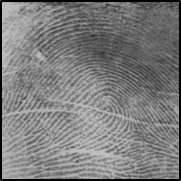}} \hfil
    \subfloat[]{\includegraphics[width=.18\linewidth]{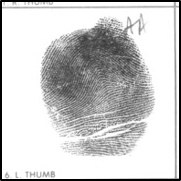}} \hfil
    \subfloat[]{\includegraphics[width=.18\linewidth]{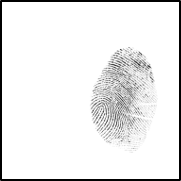}} \hfil
    \subfloat[]{\includegraphics[width=.18\linewidth]{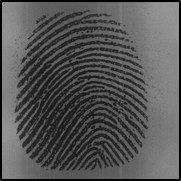}} \hfil
    \subfloat[]{\includegraphics[width=.18\linewidth]{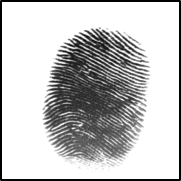}} \hfil
    \subfloat[]{\includegraphics[width=.18\linewidth]{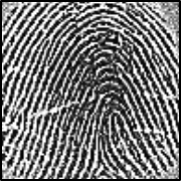}} \hfil
    \subfloat[]{\includegraphics[width=.18\linewidth]{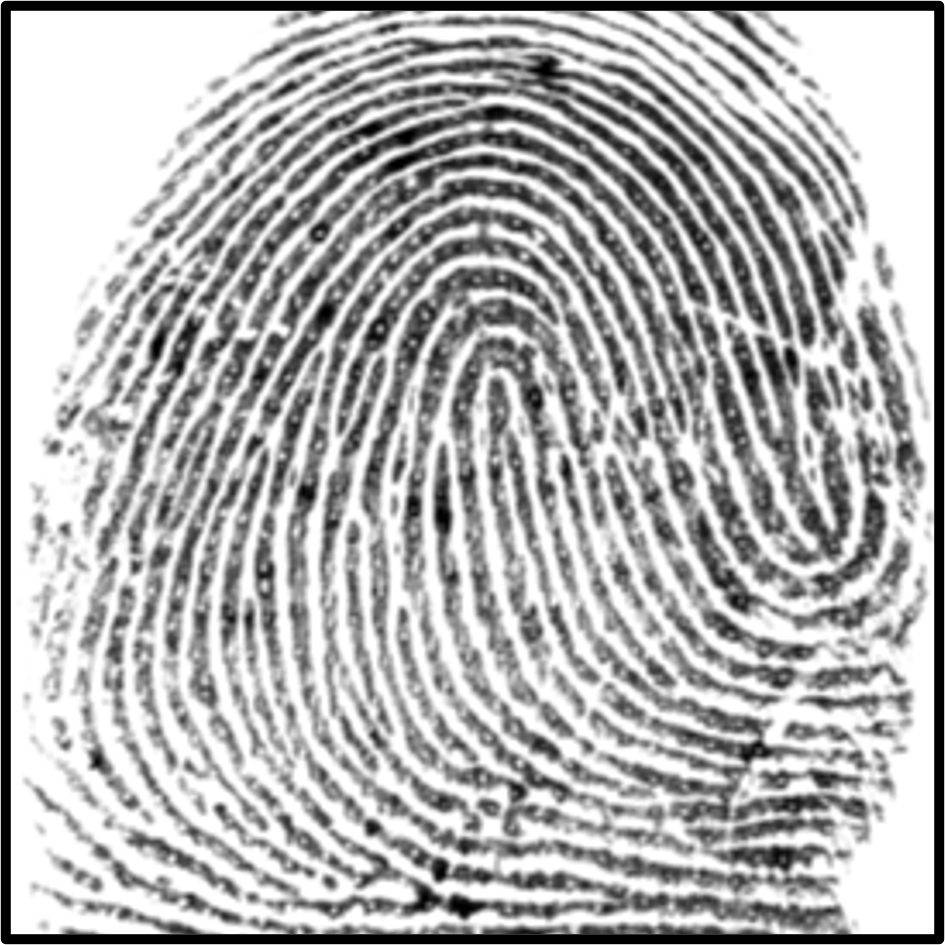}} \hfil
    \subfloat[]{\includegraphics[width=.18\linewidth]{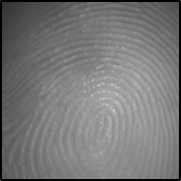}} \hfil
    \subfloat[]{\includegraphics[width=.18\linewidth]{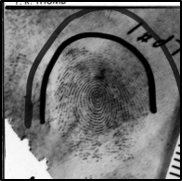}} \hfil
    \subfloat[]{\includegraphics[width=.18\linewidth]{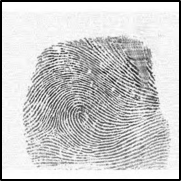}} \hfil
    \caption{Image examples from different fingerprint datasets (a) NIST SD4, (b) NIST SD14, (c) DPF, (d) FVC2002 DB3A, (e) FVC2004 DB1A, (f) FVC2006 DB1A, (g) N2N Plain, (h) PolyU CL2CB, (i) NIST SD27, (j) THU Contact10K. Images are input at 500 ppi, and they have been rescaled here to ensure optimal visualization.}
    \label{fig:datasets}
    \vspace{-0.5cm}
\end{figure}

Multiple datasets with various impression types, including rolled, plain, latent, and contactless fingerprints, are utilized for evaluation. Table \ref{tab:datasets} provides the details of all datasets and several fingerprint examples from these datasets are shown in Fig. \ref{fig:datasets}. 

\begin{table}[!t]
    \centering
    \begin{threeparttable}
      \caption{Fingerprint datasets used in experiments.}
      \label{tab:datasets}
      \vspace{-0.2cm}
      \begin{tabular}{lllcc}
        \toprule
        \textbf{Type} & \textbf{Dataset}     & \textbf{Sensor}                         & \multicolumn{1}{c}{\textbf{Quantity}}                          & \multicolumn{1}{c}{\textbf{Usage}} \\
        \midrule
        \multirow{2}{*}{Rolled}        & NIST SD4             & Inking                                  & 4,000                                 & test \\
                      & NIST SD14            & Inking                                  & 48,000                               & train \\
        \hhline
        \multirow{5}{*}{Plain} & FVC2002 DB3A     & Capacitive                      & 800                                & test \\
        & FVC2004 DB1A     & Optical               & 800                                 & test \\
        & FVC2006 DB1A     & Electric field  & 1,680                                & test \\
        & N2N Plain\tnote{a}   & Optical                                 & 4,000                      & test \\
        & DPF  & Optical                                 & 1,287       & test \\
        \hhline
        Latent        & NIST SD27\tnote{b}   & -                                       & 516                    & test \\
        \hhline
        Contactless   & PolyU CL2CB\tnote{c} & Optical/Camera                          & 2,016 & test \\
        \bottomrule
      \end{tabular} 
      \begin{tablenotes}
        \item[a] Subset R and S in NIST SD302b.
        \item[b] 10,458 fingerprints (plain or rolled) from our internal dataset THU Contact10K were added as gallery.
        \item[c] Original contactless fingerprints are scaled to match the same mean ridge period of contact-based fingerprints\cite{cui2023monocular}, and no warping was performed.
    \end{tablenotes}
\end{threeparttable}
\vspace{-0.3cm}
\end{table}

\begin{table*}[!t]
  \centering
  \scriptsize
  \caption{Verification and recognition accuracy (\%) on several fingerprint datasets. TAR@FAR=0.1\% is reported.}
  \label{tab:diff_method_comparison}
  \vspace{-0.2cm}
  \begin{threeparttable}
    \resizebox{\textwidth}{!}{%
    \begin{tabular}{lc*{1}{p{.04\linewidth}<{\centering}}c*{10}{p{.045\linewidth}<{\centering}}}
      \toprule
      \multicolumn{1}{c}{\multirow{2}{*}{\textbf{Method}\tnote{1}}}  &
      \multicolumn{1}{c}{\multirow{2}{*}{\textbf{Type}}}           &
      \multicolumn{2}{c}{\textbf{NIST SD4}}                  &
      \multicolumn{1}{c}{\textbf{FVC02 DB3}}                 &
      \multicolumn{1}{c}{\textbf{FVC04 DB1}}                 &
      \multicolumn{1}{c}{\textbf{FVC06 DB1}}                 &
      \multicolumn{2}{c}{\textbf{N2N Plain}}             &
      \multicolumn{1}{c}{\textbf{DPF}}                     &
      \multicolumn{1}{c}{\textbf{PolyU}}             &
      \multicolumn{2}{c}{\textbf{NIST SD27}}             \\  
      \cmidrule(lr){3-4} \cmidrule(lr){5-7} \cmidrule(lr){8-9} \cmidrule(lr){10-11}  \cmidrule(lr){12-13}
                                                            &  & \textbf{TAR}\tnote{2} & \textbf{Rank-1}&
                                                            \multicolumn{3}{c}{\textbf{TAR}}  & \textbf{TAR}\tnote{2}  & \textbf{Rank-1} & \multicolumn{2}{c}{\textbf{TAR}}  & \textbf{TAR}  & \textbf{Rank-1} \\
      \midrule
      VeriFinger v12.0                & Minutia-based                &  99.80 & 99.65 & \textbf{99.43} & 98.93 & 92.65 & \textbf{99.30} & 99.20 & 98.51 & \textbf{97.21} & 53.10 & 59.69 \\
      DMD\cite{pan2024latent}        & Minutia-based                               &  99.80 & 99.80 & 99.36 & 98.96 & 93.89 & 99.15 & \textbf{99.35} & \textbf{98.78} & 90.43 & 81.78 & 79.84 \\
      \hhline
      DeepPrint\cite{engelsma2021learning}\tnote{3}  &   Fixed-length             &  96.25 & 98.60 & 70.96 & 94.82 & 83.02 & 93.35 & 96.00 & 64.86 & 56.05 & 34.88 & 25.58 \\
      V-DeepPrint\cite{engelsma2021learning}\tnote{4}  &  Fixed-length              &  96.55 & 99.00 & 69.85 & 94.00 & 82.39 & 95.00 & 97.15 & 67.05 & 54.20 & 39.53 & 28.29 \\
      AFRNet\cite{grosz2023afrnet}\tnote{5}    &    Fixed-length                  &  94.75 & 97.40 & 80.25 & 98.14 & 83.31 & 96.90 & 97.40 & 77.36 & 52.61 & 45.30 & 35.66 \\
      V-AFRNet\cite{grosz2023afrnet}\tnote{6}   &   Fixed-length                  &  97.90 & 99.30 & 49.82 & 87.43 & 59.49 & 95.70 & 97.15 & 77.45 & 10.69 & 36.05 & 25.58 \\
      MultiScale\cite{gu2022latent}    &   Fixed-length          &  97.80 & 98.85 & 83.36 & 97.36 & 47.37 & 97.80 & 98.10 & 87.68 & 64.13 & 44.57 & 43.80 \\
      FDD (binary)            & Fixed-length          &  99.50 & 99.60 & 95.50 & 98.71 & 86.99 & 98.35 & 98.30 & 88.90 & 81.21 & 43.80 & 44.19 \\
      FDD                     &  Fixed-length           &  99.60 & 99.55 & 96.71 & 98.79 & 90.87 & 98.50 & 98.40 & 91.61 & 86.62 & 53.49 & 50.00 \\
      \hhline
      Fusion\tnote{7}     & --                                  & \textbf{99.85} & \textbf{99.80} & 99.39 & \textbf{99.36} & \textbf{94.33} & \textbf{99.30} & \textbf{99.35} & \textbf{98.78} & 93.62 & \textbf{82.95} & \textbf{82.56} \\
      \bottomrule     
    \end{tabular}
    }
    \begin{tablenotes}
      \item[1] All methods were trained using the same fingerprint data.
      \item[2] TAR@FAR=0.01\%
      \item[3] Performance of reimplemented DeepPrint.
      \item[4] Performance of DeepPrint reimplemented by replacing STN with the pose estimation method in \cite{duan2023estimating}.
      \item[5] Performance of reimplemented AFRNet.
      \item[6] Performance of AFRNet reimplemented by replacing STN with the pose estimation method in \cite{duan2023estimating}. 
      \item[7] The score-level fusion of FDD and DMD employs fusion weights learned from our other private fingerprint datasets. 
    \end{tablenotes}
  \end{threeparttable}
  \vspace{-0.4cm}
\end{table*}

\begin{table}[!t]
\centering
\caption{Evaluation of FDD $f$ with different feature dimension size $C$ on several challenging datasets. And $f \in \mathbb{R}^{2C \times 16 \times 16}$.}
\label{tab:dimension_reduction}
\vspace{-0.2cm}
\begin{tabular}{lc*{5}{p{.07\linewidth}<{\centering}}}
    \toprule
    \multicolumn{1}{c}{\multirow{2}{*}{\textbf{\makecell{ \# Dimension }}}}  &
    \multicolumn{5}{c}{\textbf{TAR@FAR=0.1\% (\%)}} \\  
    \cmidrule{2-6} & \multicolumn{1}{c}{\textbf{\makecell{FVC2002 \\ DB3A}}} & \multicolumn{1}{c}{\textbf{\makecell{FVC2006 \\ DB1A}}} & \multicolumn{1}{c}{\textbf{\makecell{DPF}}} & \multicolumn{1}{c}{\textbf{\makecell{PolyU \\ CL2CB}}} & \multicolumn{1}{c}{\textbf{\makecell{NIST \\ SD27}}} \\
    \midrule
    $C=1$ & 94.79 & 81.37 & 90.82 & 86.14 & 41.47 \\
    $C=3$ & 96.00 & 88.59 & 90.82 & 89.15 & 51.16 \\
    $C=6$ & 96.71 & 90.87 & 91.61 & 86.62 & 53.49\\
    $C=9$ & 94.56 & 90.50 & 90.91 & 86.40 & 56.98 \\
    $C=12$ & 96.64& 89.31 & 91.60 & 87.44 & 59.69 \\
    \bottomrule
\end{tabular}
\end{table}

\begin{table}[t]
  \vspace{-0.1cm}
  \centering
  \caption{Efficiency (seconds per sample or pair) of Different Methods.}
  \label{tab:efficiency}
  \vspace{-0.2cm}
  
  \begin{threeparttable}
      \begin{tabular}{lc*{3}{p{.22\linewidth}<{\centering}}}
          \toprule
          & \textbf{\makecell{Pose\\Estimation}} & \textbf{\makecell{Descriptor\\Extraction}} & \textbf{\makecell{Descriptor\\Matching}}\\
          \midrule 
          VeriFinger & -- & 1.98 & $5.48\times10^{-3}$ \\
          DMD & -- & $4.60\times10^{-2}$ & $1.25\times10^{-3}$ \\
          \hhline
          DeepPrint & -- & $1.47\times10^{-3}$ & $5.12\times10^{-9}$ \\
          AFRNet & -- & $1.15\times10^{-3}$ &  $2.14\times10^{-8}$ \\
          MultiScale & $2.81\times10^{-3}$ & $3.18\times10^{-3}$ & $2.63\times10^{-7}$\\
          \hhline
          FDD(binary) & $2.81\times10^{-3}$ & $6.13\times10^{-4}$ & $5.16\times10^{-8}$ \\
          FDD & $2.81\times10^{-3}$ & $6.13\times10^{-4}$ & $6.53\times10^{-8}$ \\
          \bottomrule
      \end{tabular}
      \begin{tablenotes}
          \item[*] The efficiency results are calculated based on NIST SD4.
      \end{tablenotes}
  \end{threeparttable}
  \vspace{-0.3cm}
\end{table}

We utilize the first 24,000 pairs of rolled fingerprints from NIST SD14 for training. Note that no fine-tuning is applied on any other dataset. Diverse Pose Fingerprint dataset (DPF) \cite{duan2023estimating}, which is collected from fingers with various 3D poses, consists of 143 rolled fingerprints, each accompanied by 8 plain fingerprints of various poses. For the FVC series dataset, we selected three of the most challenging ones: FVC2002 DB3A, FVC2004 DB1A, and FVC2006 DB1A.


\subsection{Compared Methods}
We compared our method with previous high-performing fixed-length fingerprint descriptors, including DeepPrint \cite{engelsma2021learning}, AFRNet \cite{grosz2023afrnet}, and MultiScale \cite{gu2022latent}. Since the code of DeepPrint and AFRNet is not available to us, we reimplemented them using the same training data to comprehensively evaluate their performance on multiple datasets. Specifically, we excluded the feature realignment strategy in AFRNet because its usage significantly increases computational complexity, thereby losing the significance of fixed-length representation. MultiScale \cite{gu2022latent} and our method employed identical pose estimation results obtained from \cite{duan2023estimating}. Additionally, DeepPrint and AFRNet were evaluated using their original STN modules and the pose estimation method from \cite{duan2023estimating}, respectively. This ensures consistency across all compared methods, emphasizing that the observed matching performance solely reflects the efficacy of the fingerprint representations. In addition, we evaluated two state-of-the-art minutiae-based descriptor methods, VeriFinger and DMD \cite{pan2024latent}, for comparison.


\begin{figure}[!t]
  \centering
  \subfloat[\label{fig:qua_nist4}]{\includegraphics[width=0.47\linewidth]{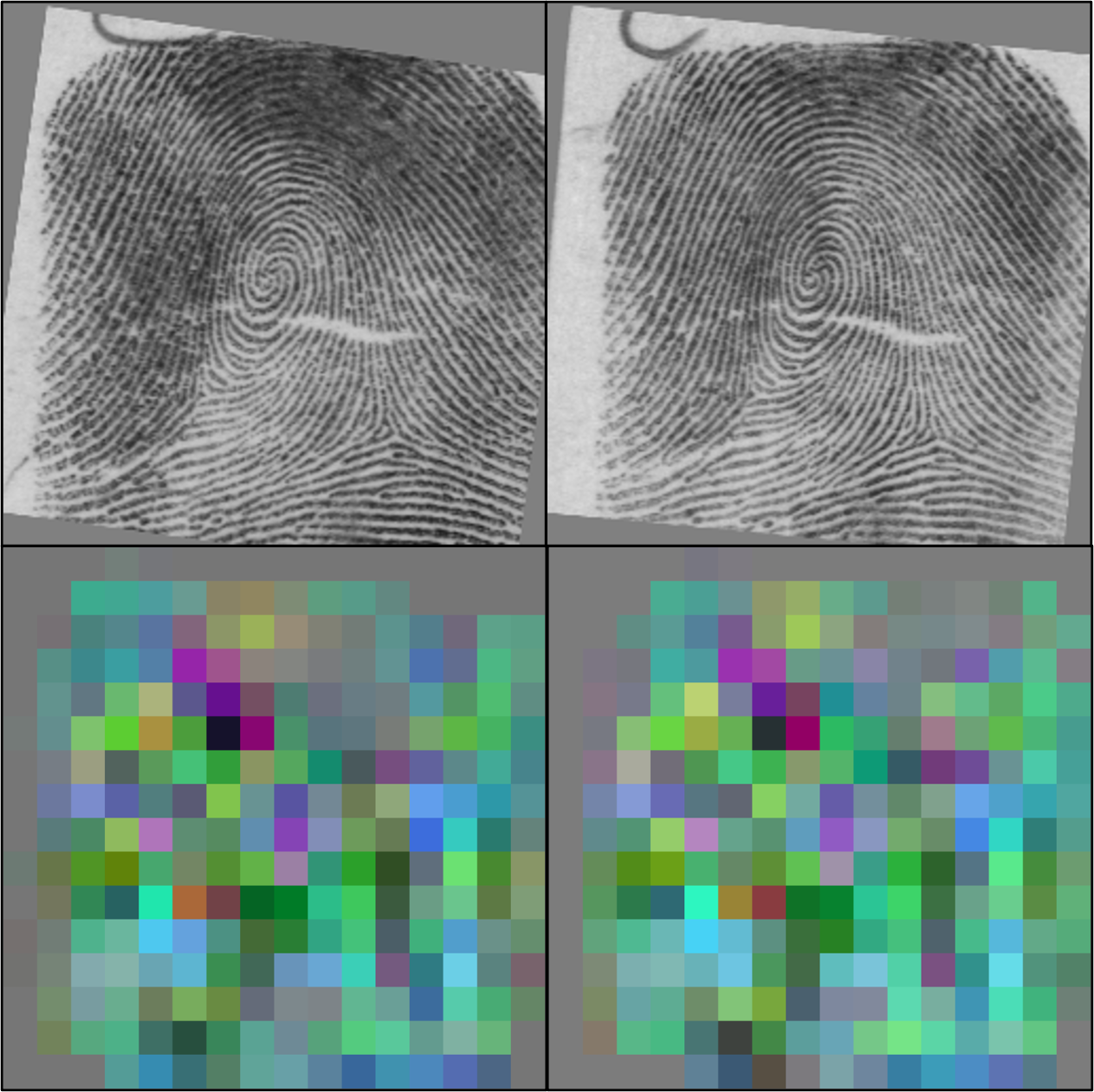}} \hfil
  \subfloat[\label{fig:qua_dpf}]{\includegraphics[width=0.47\linewidth]{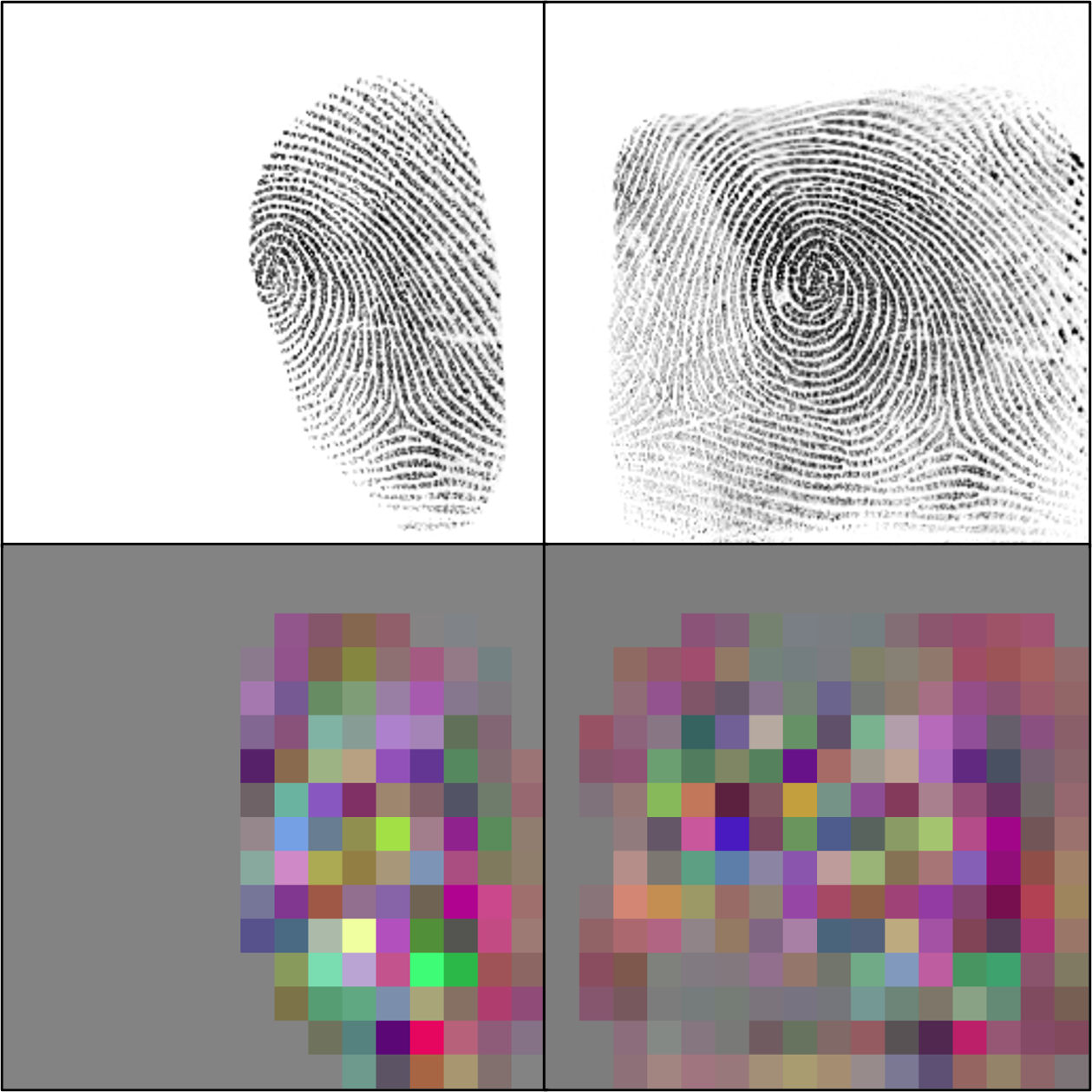}} \hfil
  \subfloat[\label{fig:qua_polyu}]{\includegraphics[width=0.47\linewidth]{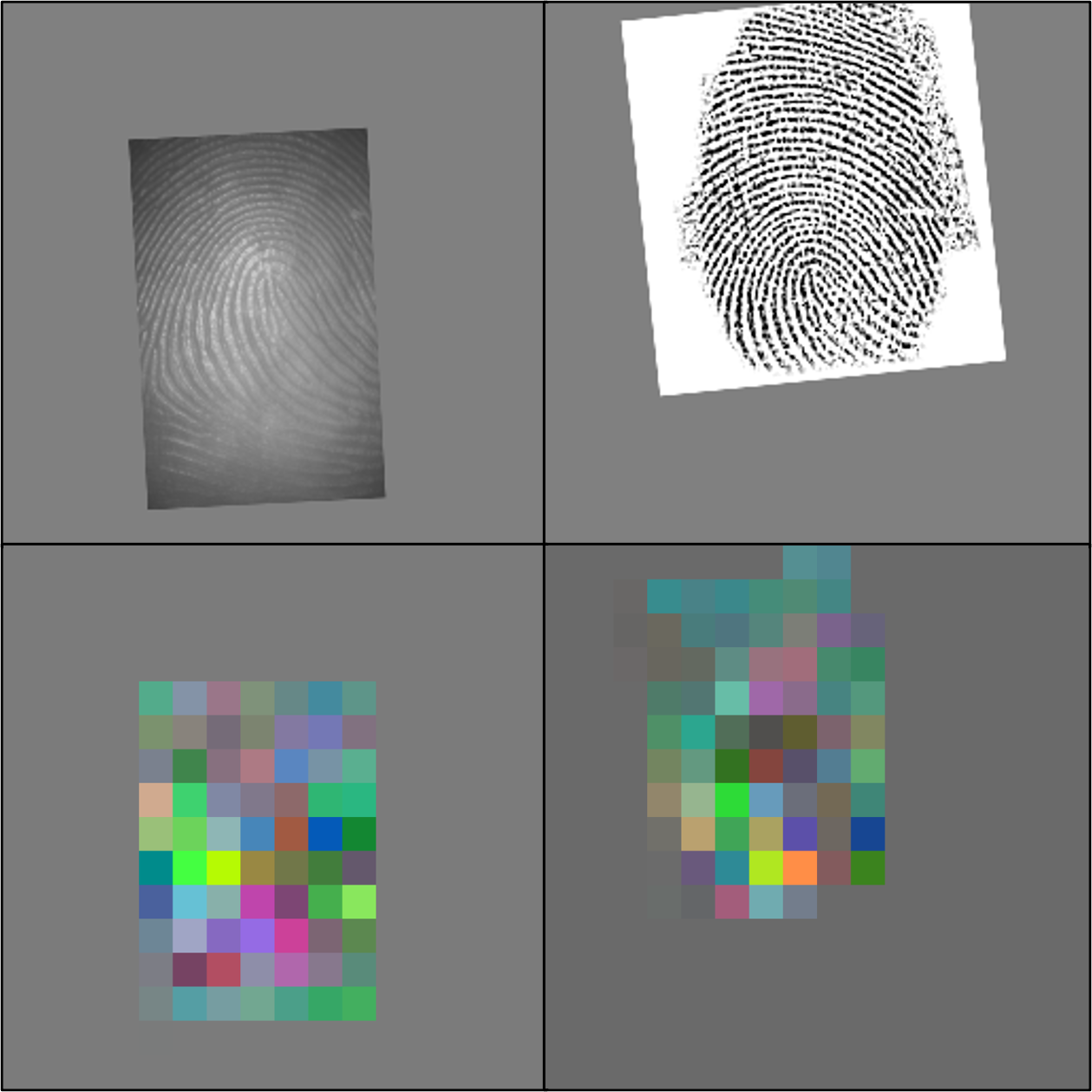}} \hfil
  \subfloat[\label{fig:qua_nist27}]{\includegraphics[width=0.47\linewidth]{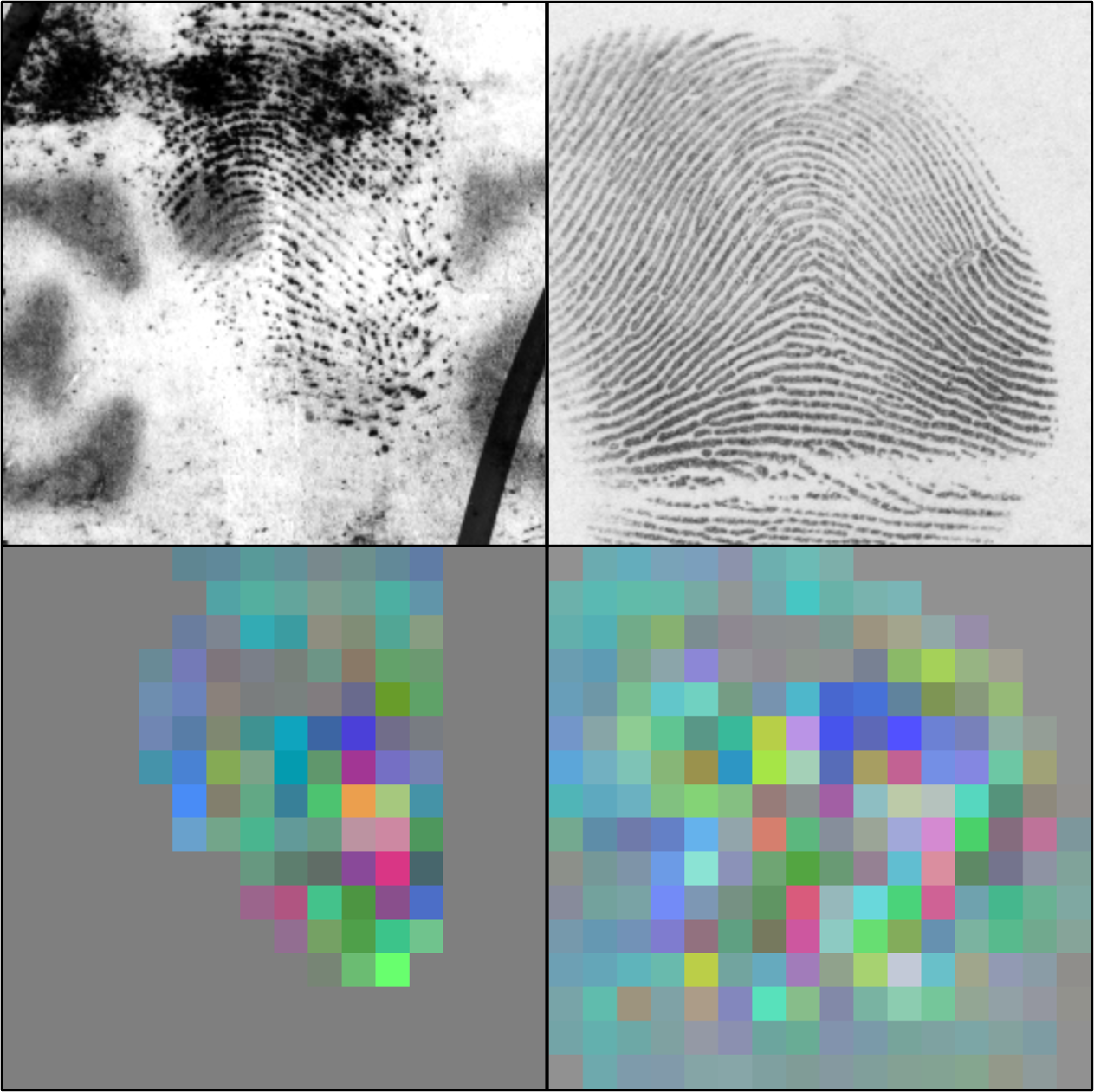}} \hfil
  \caption{Examples of the FDD extracted from genuine pairs of (a) NIST SD4, (b) DPF, (c) PolyU CL2CB, and (d) NIST SD27. The fingerprint images shown in the figure have been aligned based on their estimated poses.} 
  \label{fig:qualitative_show}
  \vspace{-0.3cm}
\end{figure}

\subsection{Results}
  As shown in Tab. \ref{tab:diff_method_comparison}, FDD demonstrates superior performance compared to the other fixed-length representation methods on all evaluated datasets, particularly on incomplete or partial fingerprints, contactless fingerprints, and latent fingerprints. Specifically, our approach outperforms the previous best performance by 16.01\% on FVC2002 DB3A, 9.07\% on FVC2006 DB1A, 35.07\% on PolyU CL2CB, and 18.07\% on NIST SD27 in terms of TAR@FAR=0.1\%. It clearly highlights the advantages of dense descriptors in matching partial fingerprints, isolating background noise, and facilitating cross-modal fingerprint matching. Additionally, we binarize the descriptor features with a threshold of 0 to obtain a binary version of FDD, and the similarity score are calculated using XOR. It still exhibits great matching performance. Although the matching performance of FDD currently falls short of minutiae-based methods, score fusion between FDD and minutia-based representation like DMD\cite{pan2024latent} yields superior results compared to either method alone, highlighting their complementarity. Furthermore, as shown in Tab. \ref{tab:efficiency}, FDD significantly outperforms minutiae-based methods in efficiency.
  
  Fig. \ref{fig:qualitative_show} shows genuine fingerprint pairs with their corresponding dense descriptors (three channels displayed). For high-quality rolled fingerprints (Fig. \ref{fig:qua_nist4}), the dense features are highly consistent. For partial fingerprints (Fig. \ref{fig:qua_dpf}), the features at corresponding positions remain correlated. In contactless-to-contact fingerprint matching (Fig. \ref{fig:qua_polyu}), the descriptors exhibit cross-modal robustness. For latent-to-rolled fingerprint matching fingerprints (Fig. \ref{fig:qua_nist27}), the descriptors effectively reduce background noise, and have high consistency in the overlapping regions.

  Furthermore, we conduct matching experiments with different representation dimensions $C$ for FDD ($f \in \mathbb{R}^{2C\times16\times16 }$). 
  In Tab. \ref{tab:dimension_reduction}, the overall trend shows that matching accuracy decreases as the feature dimension is reduced. Despite this, the low-dimensional versions of FDD still exhibit good matching performance, even surpassing other fixed-length methods on some datasets.

\section{Conclusion}

In this paper, we propose a fixed-length representation called Fixed-length Dense Descriptor (FDD), which enables efficient and interpretable fingerprint matching in large-scale datasets. 
FDD is a three-dimensional representation where the descriptor space coincides with the fingerprint space. This ensures that after aligning fingerprints using pose estimation, descriptor-level matching occurs only within overlapping foreground. To enhance representational capability, we design a dual-branch network structure to capture texture-related and minutia-related descriptors. Experiments on various fingerprint datasets show that FDD outperforms other fixed-length descriptors and its binary and low dimensional versions still achieve good matching accuracy.
However, the current extraction of dense descriptors exhibits a certain dependency on fingerprint pose accuracy. Future research will aim to develop pose-independent descriptors to enhance matching performance.
\bibliographystyle{IEEEtran}
\bibliography{IEEEabrv,refs}

\clearpage
\setcounter{page}{1}
\twocolumn[
  \begin{@twocolumnfalse}
    \begin{center}
      \textbf{\large Supplementary Material}
    \end{center}
  \end{@twocolumnfalse}
]

\begin{table}[!t]
    \centering
    \begin{threeparttable}
        \caption{Detailed Network Structure of FDD Extraction Network}
        \label{tab:FDD_Structure}
        \begin{tabularx}{\textwidth}{p{.3\linewidth}<{\centering}|p{.35\linewidth}<{\centering}|p{.35\linewidth}<{\centering}}
            \toprule
            \multicolumn{3}{c}{\textbf{Encoder}} \\
            \midrule
            \makecell[c]{Type} & \makecell[c]{Output Size \\(spatial scales, channels)} & \makecell[c]{\# of Layer} \\
            \midrule
            $7 \times 7$ Conv. Layer & $/2$, 64 & 1 \\
            \midrule
            $3 \times 3$ Residue Block & 1, 64 & 3 \\
            \midrule
            $3 \times 3$ Residue Block & $/2$, 128 & 4 \\
            \bottomrule
        \end{tabularx}
        \vspace{0.02cm}
        \begin{tabularx}{\textwidth}{p{.3\linewidth}<{\centering}|p{.35\linewidth}<{\centering}|p{.35\linewidth}<{\centering}}
            \toprule
            \multicolumn{3}{c}{\textbf{Encoder in Minutia\textbackslash Texture Branch}} \\
            \midrule
            \makecell[c]{Type} & \makecell[c]{Output Size \\(spatial scales, channels)} & \makecell[c]{\# of Layer} \\
            \midrule
            $3 \times 3$ Residue Block\tnote{a} & $/2$, 256 & 6 \\
            \midrule
            $3 \times 3$ Residue Block\tnote{b} & $/2$, 512 & 3 \\
            \bottomrule
        \end{tabularx}
        \vspace{0.02cm}
        \begin{tabularx}{\textwidth}{p{.3\linewidth}<{\centering}|p{.35\linewidth}<{\centering}|p{.35\linewidth}<{\centering}}
            \toprule
            \multicolumn{3}{c}{\textbf{Minutia Decoder}} \\
            \midrule
            \makecell[c]{Type} & \makecell[c]{Output Size \\(spatial scales, channels)} & \makecell[c]{\# of Layer} \\
            \midrule
            $3 \times 3$ Conv. Layer & 1, 128 & 6 \\
            \midrule
            $4 \times 4$ Deconv. Layer & $2$, 64 & 2 \\
            \midrule
            $3 \times 3$ Conv. Layer & 1, 6 & 1 \\
            \bottomrule
        \end{tabularx}
        \vspace{0.02cm}
        \begin{tabularx}{\textwidth}{p{.3\linewidth}<{\centering}|p{.35\linewidth}<{\centering}|p{.35\linewidth}<{\centering}}
            \toprule
            \multicolumn{3}{c}{\textbf{Mask Decoder}} \\
            \midrule
            \makecell[c]{Type} & \makecell[c]{Output Size \\(spatial scales, channels)} & \makecell[c]{\# of Layer} \\
            \midrule
            $3 \times 3$ Conv. Layer & 1, 512 & 1 \\
            \midrule
            $1 \times 1$ Conv. Layer & 1, 1 & 1 \\
            \bottomrule
        \end{tabularx}
        \vspace{0.02cm}
        \begin{tabularx}{\textwidth}{p{.3\linewidth}<{\centering}|p{.35\linewidth}<{\centering}|p{.35\linewidth}<{\centering}}
            \toprule
            \multicolumn{3}{c}{\textbf{Descriptor Decoder in Minutia\textbackslash Texture Branch }} \\
            \midrule
            \makecell[c]{Type} & \makecell[c]{Output Size \\(spatial scales, channels)} & \makecell[c]{\# of Layer} \\
            \midrule
            $3 \times 3$ Conv. Layer & 1, 512 & 1 \\
            \midrule
            $1 \times 1$ Conv. Layer & 1, 6 & 1 \\
            \bottomrule
        \end{tabularx}
        \begin{tablenotes}
            \item[a] Minutia Map Decoder connects from here.
            \item[b] The branches for Mask Decoder and Minutia\textbackslash Texture Descriptor Decoder connect from here.
        \end{tablenotes}
    \end{threeparttable}
\end{table}

\end{document}